\documentclass[conference]{IEEEtran}
\IEEEoverridecommandlockouts
\usepackage{cite}
\usepackage{amsmath,amssymb,amsfonts}
\usepackage{algorithmic}
\usepackage{graphicx}
\usepackage{textcomp}
\usepackage{xcolor}
\usepackage{color, soul}
\def\BibTeX{{\rm B\kern-.05em{\sc i\kern-.025em b}\kern-.08em
    T\kern-.1667em\lower.7ex\hbox{E}\kern-.125emX}}
\begin{document}

\title{Were You Helpful - Predicting Helpful Votes from Amazon Reviews}

\author{
\IEEEauthorblockN{Harrison Kung}
\IEEEauthorblockA{\textit{Computer Engineering} \\
\textit{University of California, San Diego}\\
La Jolla, USA \\
hkung@ucsd.edu}
\and
\IEEEauthorblockN{Dominic Orlando}
\IEEEauthorblockA{\textit{Electrical Engineering} \\
\textit{University of California, San Diego}\\
La Jolla, USA \\
dorlando@ucsd.edu}
\and
\IEEEauthorblockN{Emin Kirimlioglu}
\IEEEauthorblockA{\textit{Computer Science} \\
\textit{University of California, San Diego}\\
La Jolla, USA \\
ekirimli@ucsd.edu}
}

\maketitle


\begin{abstract}
This project investigates factors that influence the perceived helpfulness of Amazon product reviews through machine learning techniques. After extensive feature analysis and correlation testing, we identified key metadata characteristics that serve as strong predictors of review helpfulness. While we initially explored natural language processing approaches using TextBlob for sentiment analysis, our final model focuses on metadata features that demonstrated more significant correlations, including the number of images per review, reviewer's historical helpful votes, and temporal aspects of the review.

The data pipeline encompasses careful preprocessing and feature standardization steps to prepare the input for model training. Through systematic evaluation of different feature combinations, we discovered that metadata elements we choose using a threshold provide reliable signals when combined for predicting how helpful other Amazon users will find a review. This insight suggests that contextual and user-behavioral factors may be more indicative of review helpfulness than the linguistic content itself.

\end{abstract}


\section{Introduction}
\begin{figure}
    \centering
    \includegraphics[width=\linewidth]{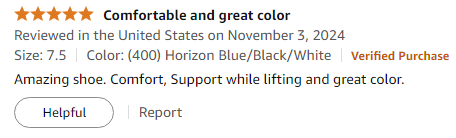}
    \caption{Amazon Review with "Helpful" button to allow customer to vote on if a review is helpful.}
    \label{fig:amazonreviewexample}
\end{figure}

In the growing e-commerce landscape, online product reviews have become important in providing information for consumers on the quality of a product. However, with the growth of user-generated content, identifying helpful reviews from the vast quantity of lower quality submissions poses a significant challenge. Amazon has implemented a system where users can vote if a review has been helpful as seen in Figure \ref{fig:amazonreviewexample}. Using this data, this paper presents a machine learning approach to automatically predict whether an Amazon product review will ever be voted as helpful. This would be useful in determining whether to promote certain reviews to the forefront, without requiring other shoppers to vote for a review which may be especially useful for new products or those with few reviews.

Our methodology tested out multiple features including natural language processing techniques by leveraging TextBlob for sentiment features (e.g. polarity and subjectivity). Other metadata features like the images per review, average rating, and the review rating were also tested to see their correlations. From there, we are able to select the features that worked to best in order to develop a framework for review helpfulness assessment utilizing multiple neural network architectures and baselines involving linear and logistic regression. By testing out MLP at different depths, RNN, Transformer, as well as our baseline models, we were able to tune our model to create a maximum accuracy of 0.9691 with MLP-64 Deep and AdamW optimization.

The key contributions of this work include:
\begin{itemize}
    \item Insights into the relationship between different features and expected helpfulness
    \item A feature engineering approach combining sentiment analysis with review metadata
    \item Correlation analysis to decide which features to keep in the final model
    \item An implementation of two baseline linear and logistic models utilizing review length 
    \item A comparison and analysis of multiple neural network models architectures (MLP, RNN, Transformer)
    \item Empirical evaluation on the Amazon Reviews 2023 dataset and results through accuracy and MSE
\end{itemize}

Our paper aims to enhance the user experience in e-commerce platforms by predicting which reviews will be deemed helpful to others.

\section{Related Works \& Literature}

\subsection{Bridging Language and Items for Retrieval and Recommendation}

Bridging Language and Items for Retrieval and Recommendation (BLAIR) leverages a learning approach to enhance retrieval and recommendation tasks by establishing connections between review text and item metadata \cite{Hou2024}. While BLAIR focuses on learning correlations between item metadata and natural language contexts, our work takes a different direction by investigating specific factors that predict review helpfulness. Through extensive feature analysis and correlation testing, we identified three key predictive features: the reviewer's historical helpful votes, number of images included in the review, and the review timestamp.

\begin{figure}
    \centering
    \includegraphics[width=\linewidth]{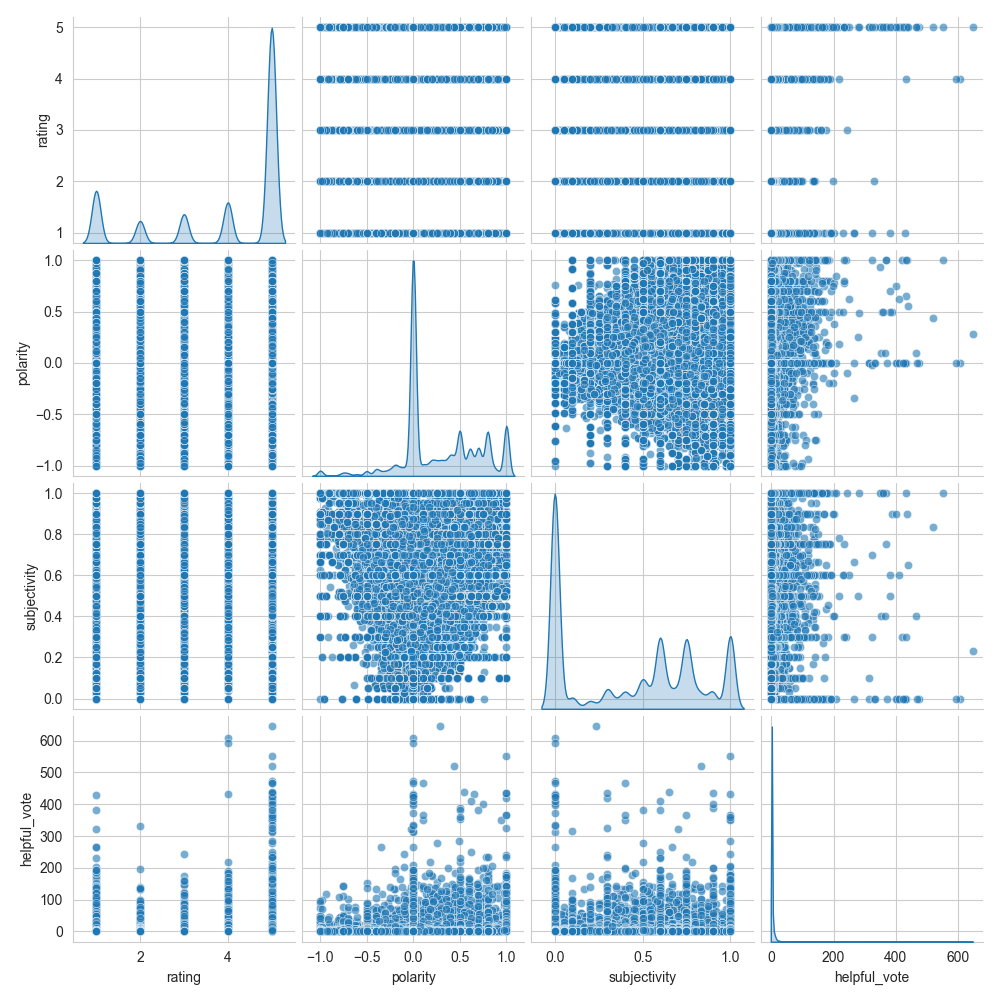}
    \caption{Scatter Matrix of Sentiment Analysis Features}
    \label{fig:scatterofsentiment}
\end{figure}

\begin{figure}
    \centering
    \includegraphics[width=\linewidth]{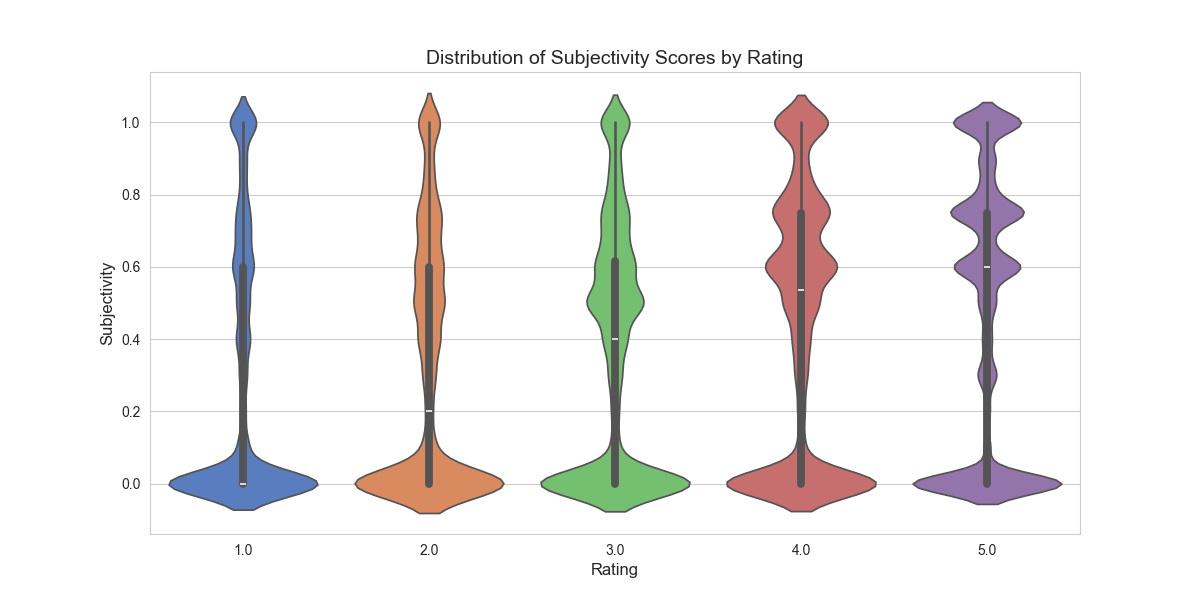}
    \caption{Violin plot of the distribution of subjectivity scores across different ratings (1.0 to 5.0)}
    \label{fig:sentimentvsratings}
\end{figure}

We initially explored sentiment analysis features such as polarity subjectivity against the helpful vote but we found low correlation \ref{fig:scatterofsentiment}, \ref{fig:sentimentvsratings}. Our rigorous feature selection process removed the sentiment analysis features from the final set of features after it was revealed that these three metadata characteristics demonstrated the strongest correlations with review helpfulness. This finding suggests that user behavior patterns and review presentation elements may be more indicative of helpfulness than linguistic or sentiment-based features. The contrast with BLAIR's approach highlights how different aspects of review data can be leveraged for distinct objectives in recommendation systems.

\subsection{Predicting Helpfulness Ratings of Amazon Product Reviews}

\cite{Rodak2024} developed a binary classifier to predict the helpfulness of Amazon electronics reviews using traditional machine learning approaches like Naive Bayes and SVMs. They incorporated various features:

\begin{itemize}
    \item Anatomical features (review length, sentence count)
    \item Metadata features (star ratings, deviation from average rating) 
    \item Lexical features (tf-idf weighted unigrams, readability metrics)
\end{itemize}

Their approach achieved 76.6\% accuracy. While our initial research direction aimed to expand on this work through neural networks and advanced sentiment analysis, our feature correlation analysis revealed that simpler metadata features were actually more predictive of review helpfulness. Specifically, we found that just three key features:

\begin{itemize}
    \item User's average helpful votes
    \item Number of images per review
    \item Review timestamp
\end{itemize}

showed the strongest correlations with helpfulness ratings, suggesting that user reputation and review presentation elements may be more reliable indicators than complex linguistic features.

\subsection{Review Popularity and Review Helpfulness: A Model for User
 Review Effectiveness}

\cite{Wu2017} proposes a conceptual framework for review effectiveness that distinguishes between review popularity and helpfulness, analyzing their relationship using Amazon data. Their research highlights that review popularity is equally important as helpfulness in evaluating review effectiveness, with certain determinants like valence having contrasting effects on each metric.

While their work focused on item retrieval and recommendation, their extensive dataset and analysis of metadata-language relationships provided valuable insights. However, our investigation took a different direction - after extensive feature analysis, we found that three key metadata features demonstrated the strongest correlations with review helpfulness:

\begin{itemize}
    \item User's average helpful votes
    \item Number of images per review
    \item Review timestamp
\end{itemize}

Although we initially planned to investigate sentiment features through TextBlob analysis, our correlation studies revealed that these simpler metadata characteristics were more reliable predictors of review helpfulness than sentiment-based features.

\section{Exploratory Analysis}
\begin{table}[htbp]
\caption{Average Number of Helpful Votes}
\begin{center}
\begin{tabular}{|c|c|c|c|c|c|}
\hline
 & \textbf{1 Star} & \textbf{2 Star} & \textbf{3 Star} & \textbf{4 Star} & \textbf{5 Star} \\
\hline
\textbf{Total \# Reviews} & 102080 & 43034 & 56307 & 79381 & 420726 \\
\textbf{Avg. Votes} & 0.964 & 0.748 & 0.732 & 0.931 & 0.956 \\
\hline
\end{tabular}
\label{tab:topic-frequency}
\end{center}
\end{table}

This project utilizes a dataset which consists of beauty products reviews on Amazon and their corresponding item metadata. Each review contains an ASIN which is an Amazon Standard Identification Number which is a unique product identifier to link it to the item's metadata. Features include the review's title and text which will be useful for sentiment analysis as well as the user's rating from one to five, the number of helpful votes that the review has received, and a list of associated images, if any.

As demonstrated in Table \ref{tab:topic-frequency}, there is a substantially higher number of 5 star ratings, with 1 star ratings being the second highest. This implies that users might be more likely to write a review for something that worked very well or not at all versus items that worked an expected or mediocre amount. Also, we can see that the average number of helpful votes is below 1 regardless of rating, but reviews of 1, 4 and 5 stars have a substantially higher average \# of helpful votes. From this we can infer that Amazon users are more likely to view the extremes when it comes to gathering data from reviews to influence their purchasing decisions. 

\begin{figure}
    \centering
    \includegraphics[width=\linewidth]{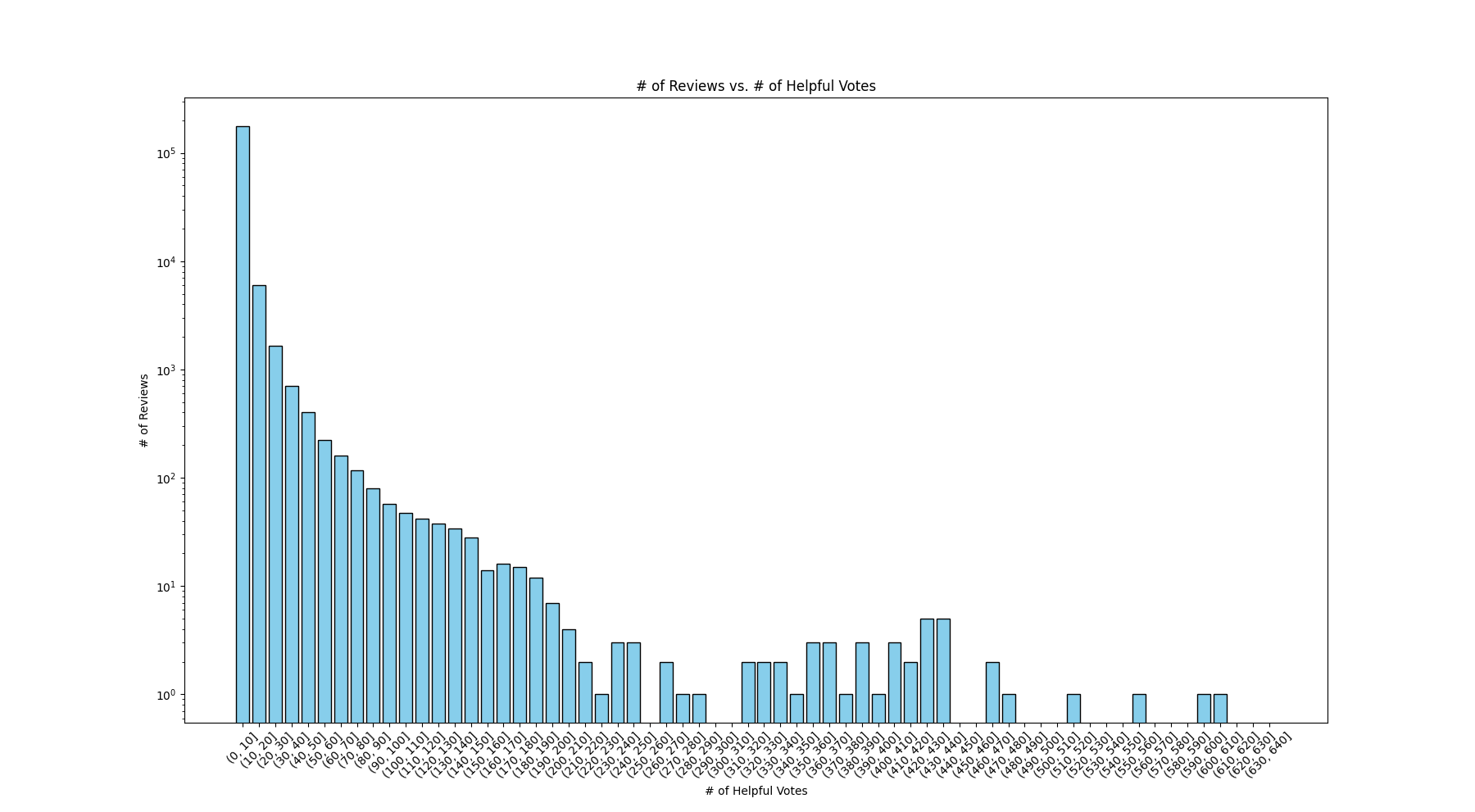}
    \caption{Histogram of the number of reviews verses the number of helpful votes received}
    \label{fig:numreviewshelpful}
\end{figure}

We also investigated the distribution of helpful votes, as shown in Figure \ref{fig:numreviewshelpful}. Evidently, very few reviews get more than 220 helpful votes, with the majority of reviews in the [0, 10] vote range. This demonstrates that the dataset is skewed towards reviews with few, if any, helpful votes, and is important to recognize when determining what to predict.

\begin{figure}
    \centering
    \includegraphics[width=\linewidth]{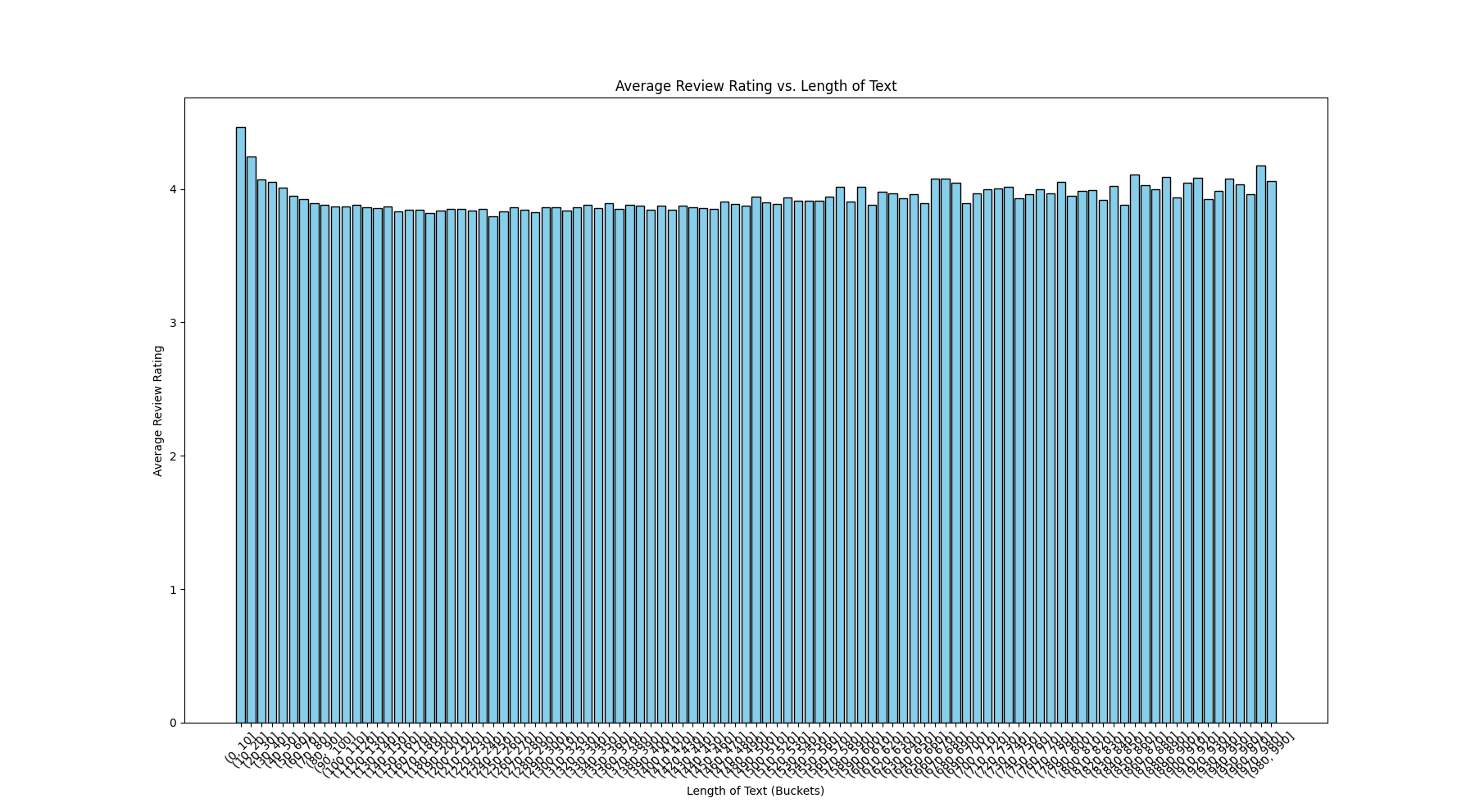}
    \caption{Histogram of the average review rating based on review text length from [0, 1000]}
    \label{fig:textlength1000}
\end{figure}

\begin{figure}
    \centering
    \includegraphics[width=\linewidth]{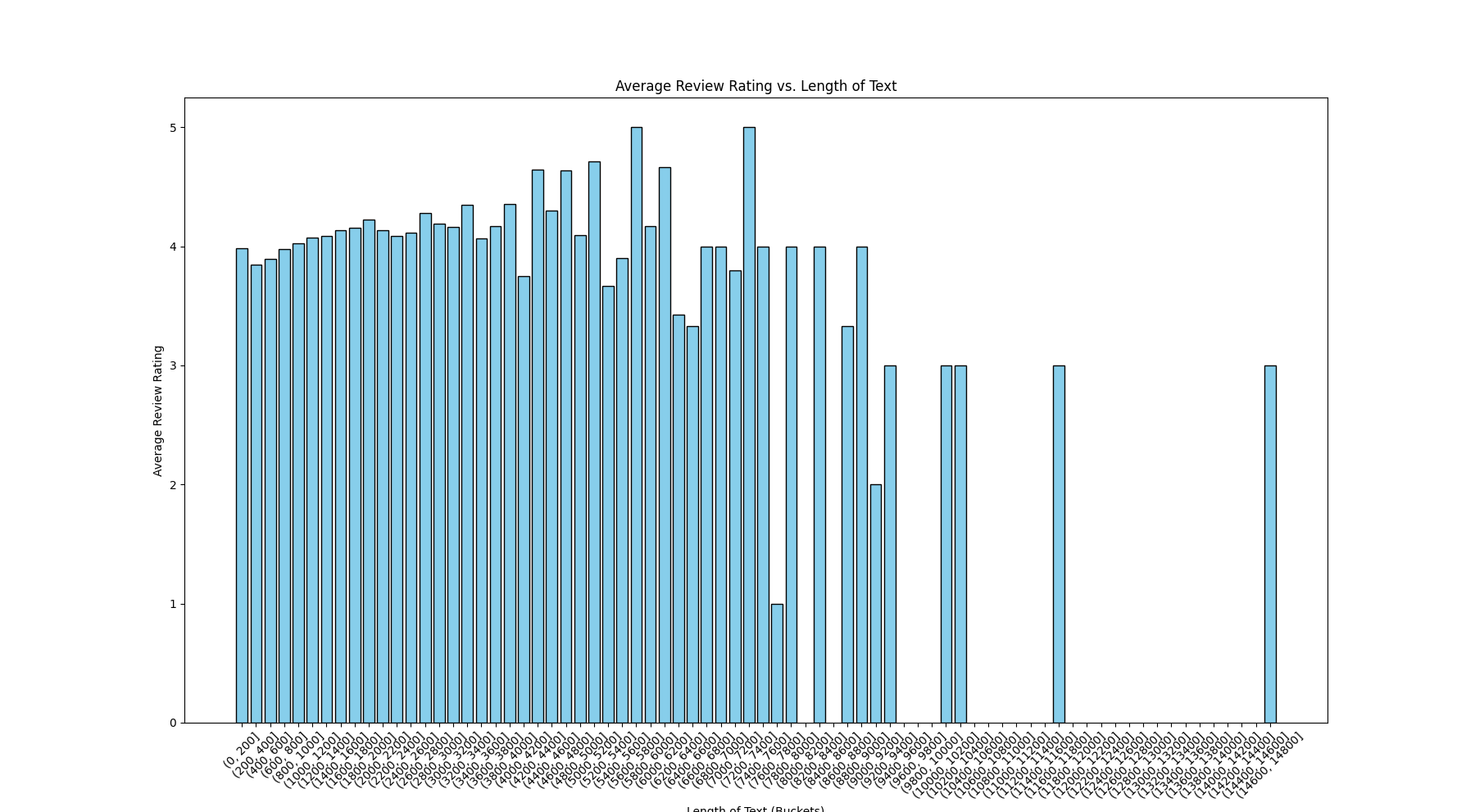}
    \caption{Histogram of the average review rating based on review text length from [0, 15000] (includes maximum length)}
    \label{fig:textlengthfull}
\end{figure}

\begin{figure}
    \centering
    \includegraphics[width=\linewidth]{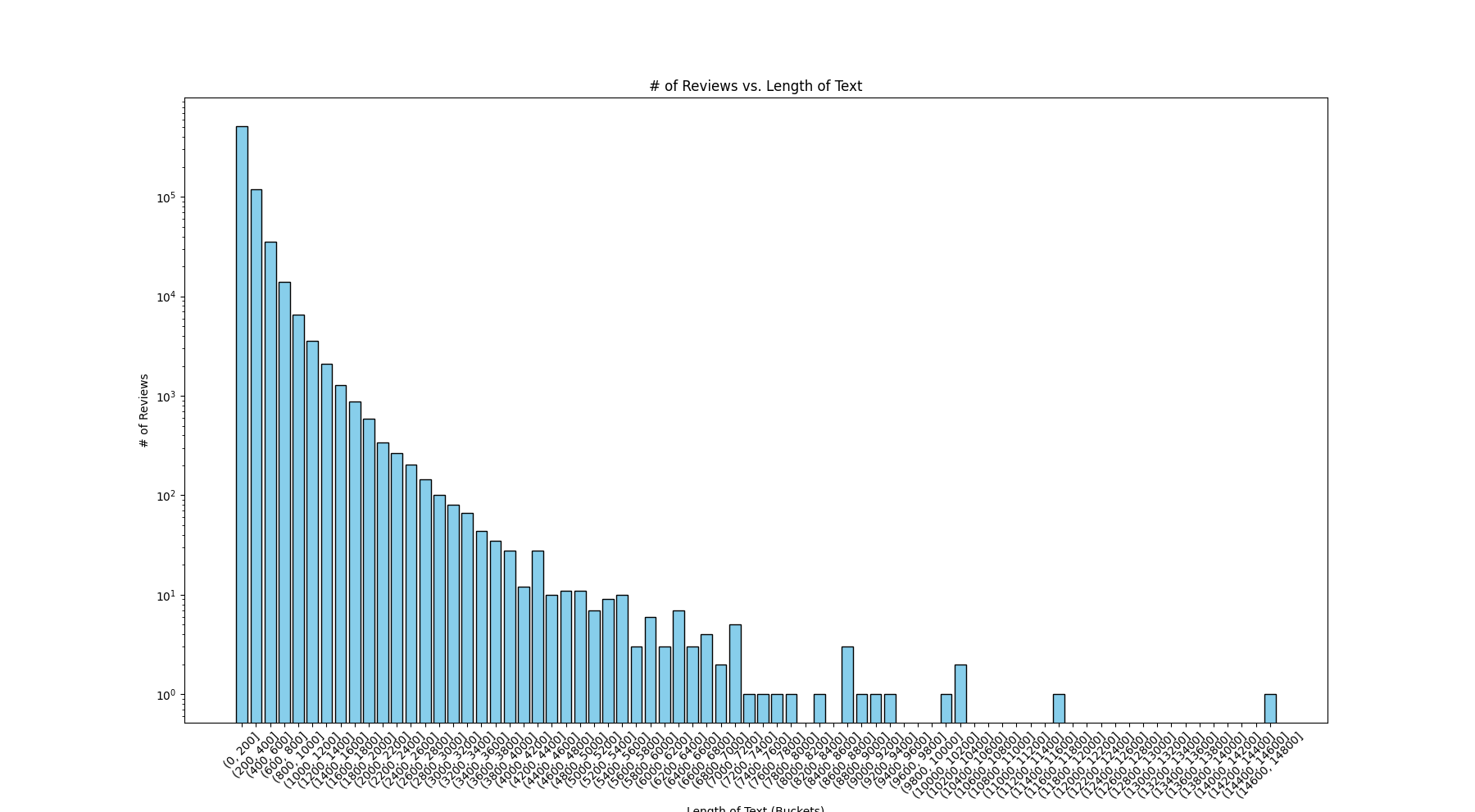}
    \caption{Histogram of logarithmic number of reviews of each review text length}
    \label{fig:logreviewcount}
\end{figure}

When exploring the relationship between the length of a review's text and the rating received, it is apparent there is a weakly positive correlation between the text length and the rating beyond the first few buckets in Figure \ref{fig:textlength1000} which encompass review length from 0 to around 70. When utilizing the entire review dataset, we can see that the correlation drops off as the review length gets longer according to Figure \ref{fig:textlengthfull}. To determine the cause of this, we then looked at the distribution of the review text lengths. As demonstrated in Figure \ref{fig:logreviewcount}, the vast majority of reviews have a text length in between [0, 200], in fact, over 500,000 of the ~701.5K total reviews have a text length in this bucket \cite{Hou2024}. The distribution of the ratings verses the text length becomes more explainable when considering that above 6,800 words there is no more than 10 reviews in any given bucket. Due to this property of the dataset, we will consider removing reviews that have a text length above 6,800 in an attempt to reduce the overall computation needed to train our model.

\begin{figure}
    \centering
    \includegraphics[width=\linewidth]{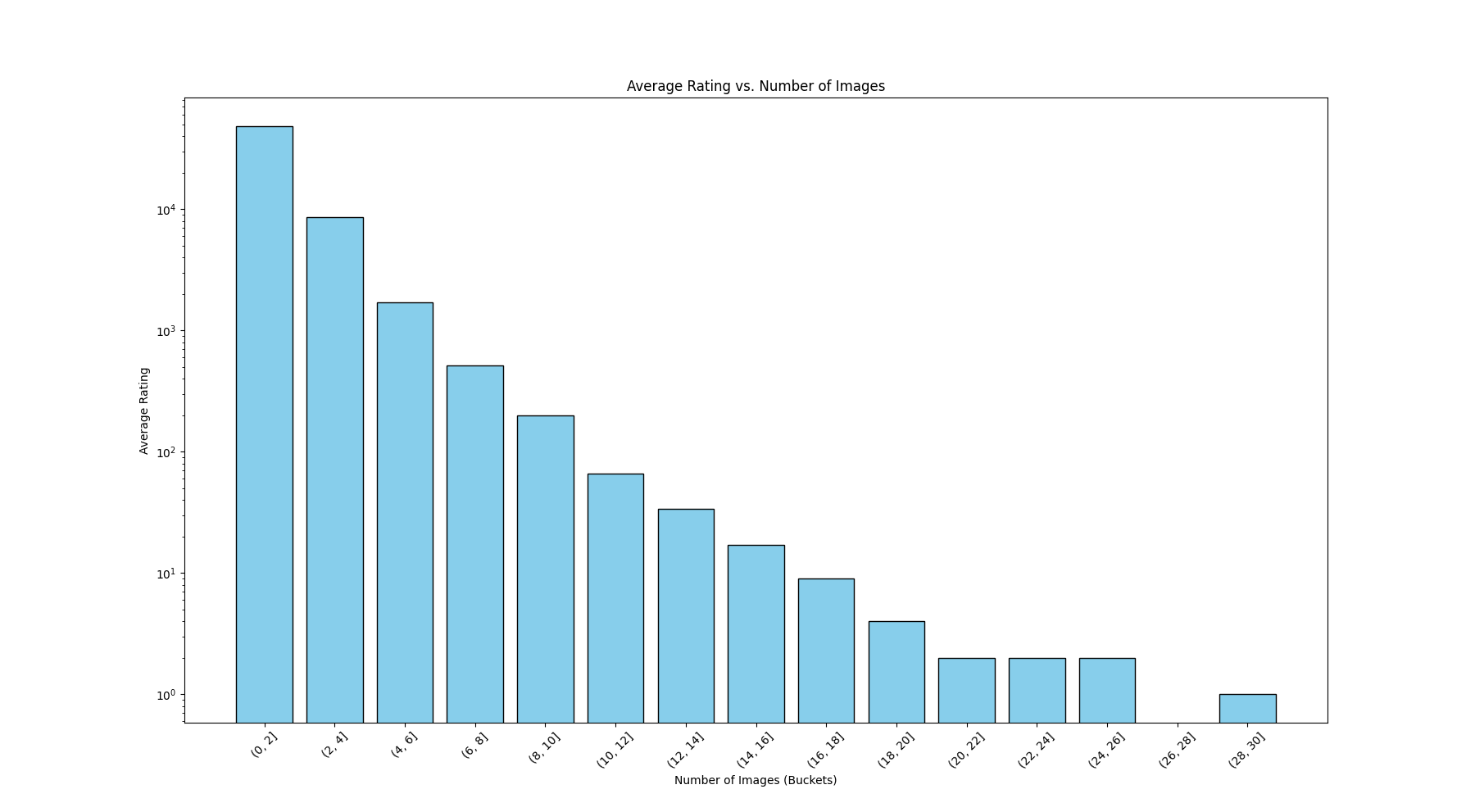}
    \caption{Histogram of logarithmic number of reviews compared to number of images in a review}
    \label{fig:logimagecount}
\end{figure}

\begin{figure}
    \centering
    \includegraphics[width=\linewidth]{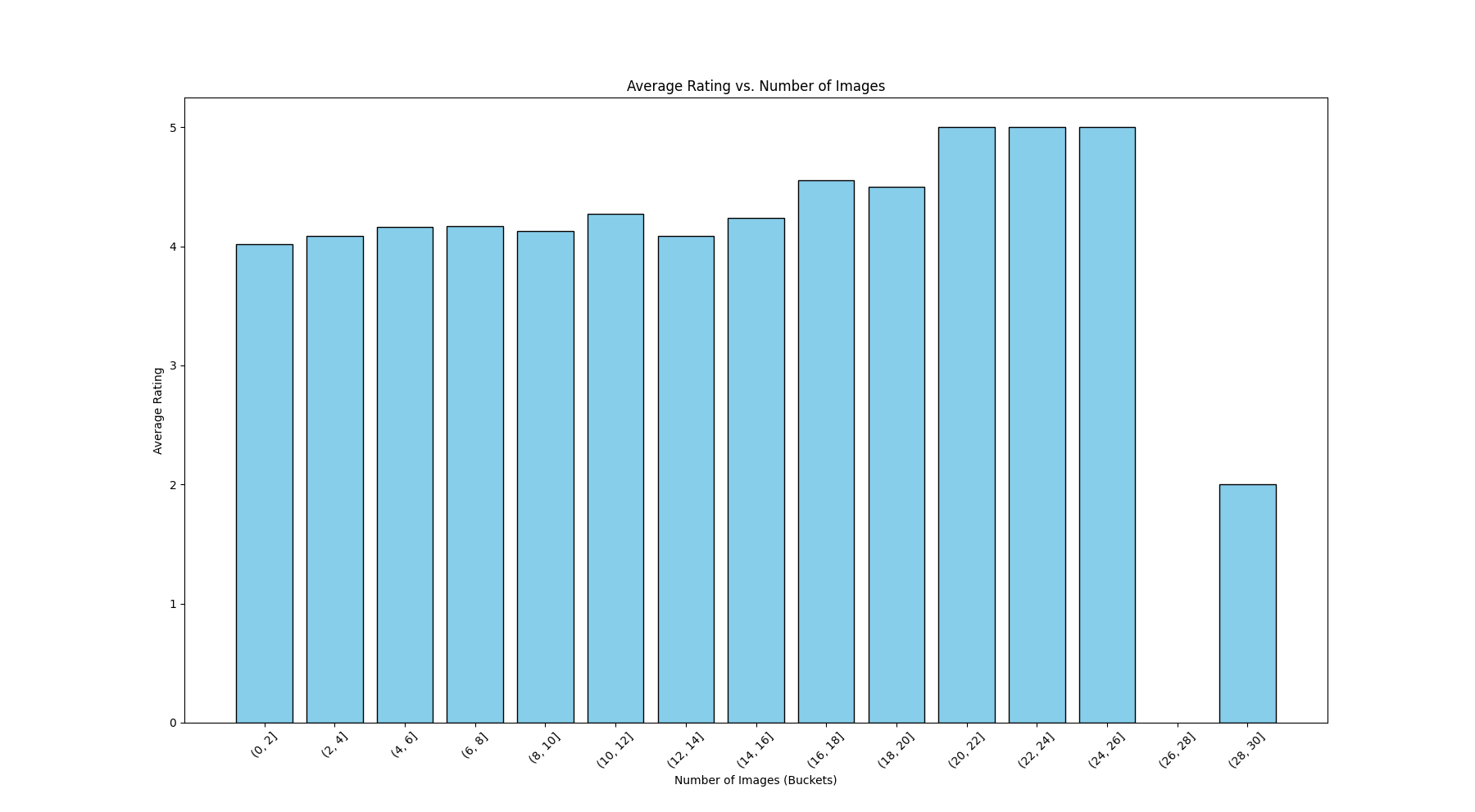}
    \caption{Average rating given compared to number of images in the review}
    \label{fig:avgratingnumimages}
\end{figure}

\begin{figure}
    \centering
    \includegraphics[width=\linewidth]{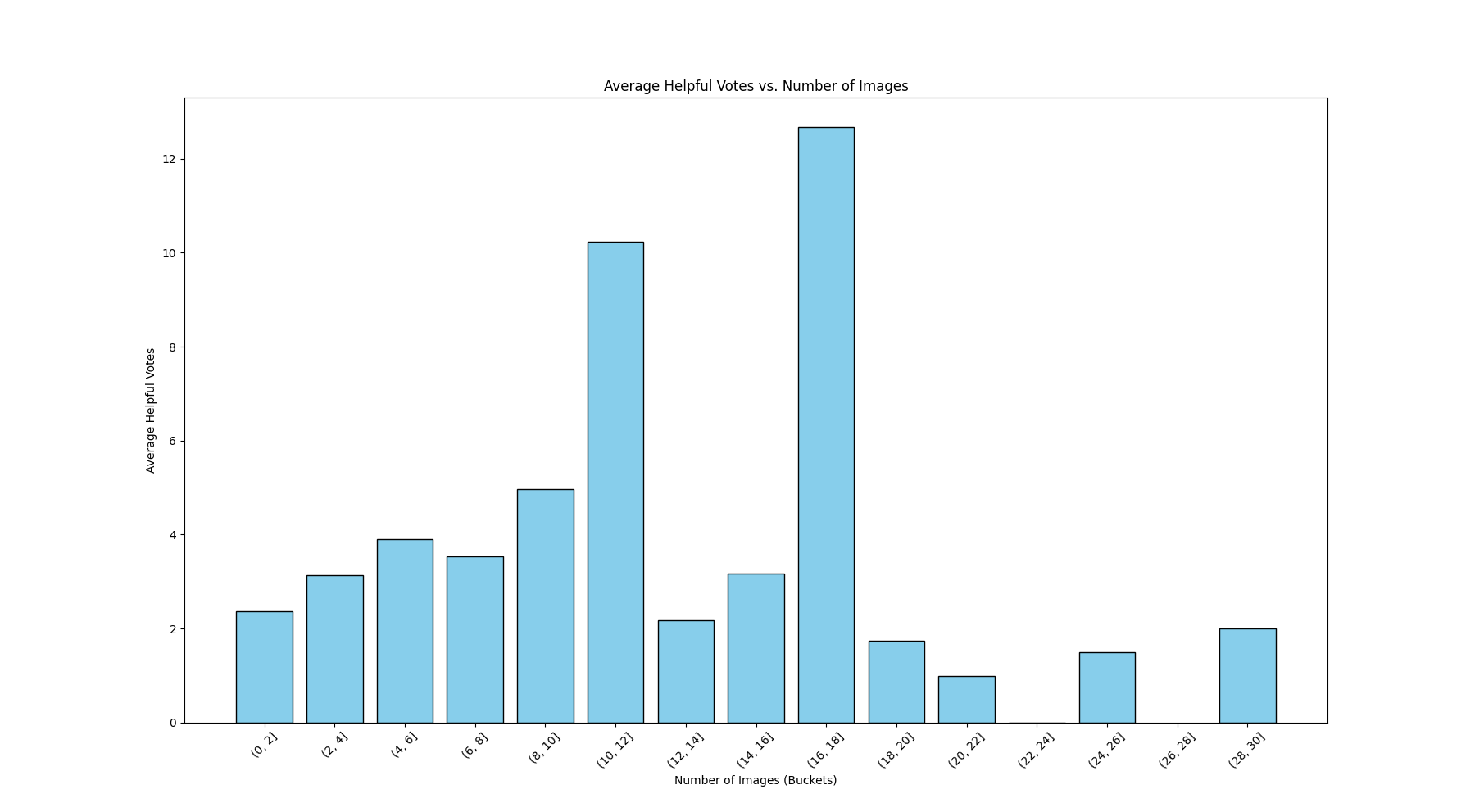}
    \caption{Average number of helpful votes given compared to number of images in the review}
    \label{fig:avghelpfulnumimages}
\end{figure}

We also consider the number of images a review contains when determining if a review is considered helpful to customers. First, we looked at the distribution of reviews based on the number of images included to get a sense of how the number of images influences both the average rating and the number of helpful votes. According to Figure \ref{fig:logimagecount} we can see that the vast majority of reviews again have 0-2 images, which is similar to the length of the review text. In this dataset, most of the reviews are relatively short and contain no images. With this in mind, we wanted to investigate the relationship between the average rating given and the number of images in reviews. As seen in Figure \ref{fig:avgratingnumimages}, there appears to be a slight positive correlation between the number of images in a review and the average rating given. When investigating these correlations, it is important to remember that there are substantially less reviews with a high number of images, so the average rating has less data to support it in the higher bins. In addition to the average rating given, we also investigated the impact the number of images a review has on how helpful the review is perceived. Evident in Figure \ref{fig:avghelpfulnumimages}, the distribution of the average reviews is multi-modal, with two modes in close proximity at [10, 12] and [16, 18]. This implies that reviews with more images are perceived as more helpful by other users.

\begin{figure}
    \centering
    \includegraphics[width=\linewidth]{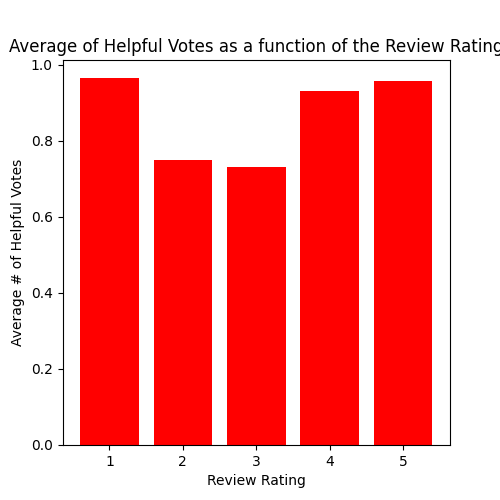}
    \caption{Average of Helpful Votes as Function of Review Ratings}
    \label{fig:averagehelpfulvotes}
\end{figure}

\begin{figure}
    \centering
    \includegraphics[width=\linewidth]{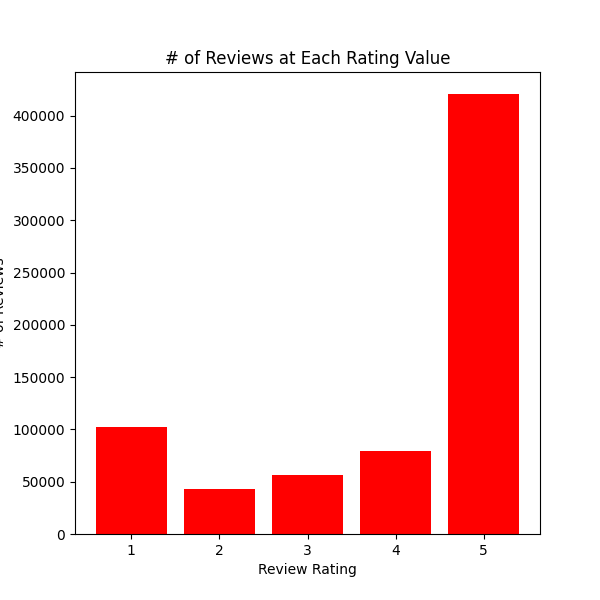}
    \caption{Number of Reviews at Each Rating Value}
    \label{fig:numberofreviews}
\end{figure}

From our exploratory investigation of our dataset, we have verified a few key assumptions. Firstly, a review is voted as helpful more when there are images in the review, and the average rating is slightly higher with more images. Secondly, the majority of ratings are 5 star, with the remaining portion dominated by 1 star ratings as seen in Figure \ref{fig:numberofreviews}. This is important to remember, as the majority of rating reviews for the specific features are high, mostly at or above 4 stars in Figure \ref{fig:averagehelpfulvotes}.

\section{Identify Predictive Task}
The predictive task we have identified is to predict if a review will be considered helpful by Amazon customers based on the content from the review and metadata of the reviewed product. Through our exploration of the dataset we choose as our features what we believe to have the greatest influence on the helpfulness of a given review.

We engineered our initial set of features by acquiring the length of the review text, calculating the polarity and subjectivity of the review text, the average rating the product received, the individual review rating, the review's number of helpful votes, and the review author's average helpful score which is computed over all reviews a user has written in the dataset. To classify if a review is considered helpful, we determined that if a review has at least 1 helpful vote, it will be considered helpful. We decided to approach the problem from more of a binary classification perspective, as our dataset is heavily skewed towards examples with few amounts of helpful votes, and also allows us to use vastly different models to compare performance. 

To assess our model's performance, we implemented several baseline approaches for comparison. Our evaluation framework included:

\begin{itemize}
    \item Linear Regression: Used to establish a basic continuous prediction baseline
    \item Logistic Regression: Implemented for binary classification comparison
\end{itemize}

For evaluating baseline performance, we employed different metrics based on the model type:

\begin{itemize}
    \item Regression Models: Mean Squared Error (MSE) to measure prediction accuracy
    \item Neural Network: Binary classification accuracy (predicting whether a review receives at least one helpful vote)
\end{itemize}

The binary nature of our neural network's predictions (0 or 1) made classification accuracy a more appropriate metric than MSE for evaluating its performance. This approach allowed us to directly compare the effectiveness of our selected features across different model architectures.

\section{Models Compared}
We evaluated several approaches for predicting review helpfulness, ranging from simple baseline models to more complex neural architectures.

\subsection{Baseline Models}
\subsubsection{Linear Regression}
Our basic linear regression model serves as a continuous prediction baseline:
\begin{equation}
\hat{y} = Wx + b
\end{equation}
where $W \in \mathbb{R}^{1 \times d}$ represents the weights for our three key features ($d=3$).

\subsubsection{Logistic Regression}
For binary classification comparison, we implemented logistic regression:
\begin{equation}
\hat{y} = \sigma(Wx + b) = \frac{1}{1+e^{-(Wx + b)}}
\end{equation}
where $\sigma$ is the sigmoid activation function.

\subsection{Neural Network Architectures}
\subsubsection{MLP-64}
MLP-64 is a feed-forward neural network for binary classification with three fully-connected layers, using dimensionality reduction and dropout regularization. The network is defined as:
\begin{equation}
\begin{split}
h_1 &= \text{Dropout}_{0.2}(\text{ReLU}(W_1x + b_1)) \\
h_2 &= \text{Dropout}_{0.2}(\text{ReLU}(W_2h_1 + b_2)) \\
\hat{y} &= \sigma(W_3h_2 + b_3)
\end{split}
\end{equation}
where $W_1 \in \mathbb{R}^{64 \times d}$, $W_2 \in \mathbb{R}^{32 \times 64}$, and $W_3 \in \mathbb{R}^{1 \times 32}$ are learnable weights, and $d$ is input dimension. 

\subsubsection{MLP-128}
This is a simpler feed-forward neural network with two fully-connected layers:
\begin{equation}
\begin{split}
h_1 &= \text{Dropout}_{0.2}(\text{ReLU}(W_1x + b_1)) \\
\hat{y} &= \sigma(W_2h_1 + b_2)
\end{split}
\end{equation}
where $W_1 \in \mathbb{R}^{128 \times d}$ and $W_2 \in \mathbb{R}^{1 \times 128}$.

\subsubsection{MLP-64-deep}
A deeper variant of MLP-64 with additional hidden layers:
\begin{equation}
\begin{split}
h_1 &= \text{Dropout}_{0.2}(\text{ReLU}(W_1x + b_1)) \\
h_2 &= \text{Dropout}_{0.2}(\text{ReLU}(W_2h_1 + b_2)) \\
h_3 &= \text{Dropout}_{0.2}(\text{ReLU}(W_3h_2 + b_3)) \\
h_4 &= \text{Dropout}_{0.2}(\text{ReLU}(W_4h_3 + b_4)) \\
\hat{y} &= \sigma(W_5h_4 + b_5)
\end{split}
\end{equation}
where $W_1 \in \mathbb{R}^{64 \times d}$, $W_2 \in \mathbb{R}^{32 \times 64}$, $W_3,W_4 \in \mathbb{R}^{32 \times 32}$, and $W_5 \in \mathbb{R}^{1 \times 32}$.

All models were trained on our three key features: user's average helpful votes, number of images per review, and review timestamp. For consistent evaluation across models, we standardized our metrics to use binary classification accuracy, determining whether a review receives at least one helpful vote (1) or none (0).

\subsection{Optimization}
Initial training of all models utilized the Adam optimizer \cite{kingma2017adammethodstochasticoptimization}, optimizing binary cross-entropy loss:
\begin{equation}
\mathcal{L}(y, \hat{y}) = -\frac{1}{N}\sum_{i=1}^N [y_i\log(\hat{y_i}) + (1-y_i)\log(1-\hat{y_i})]
\end{equation} 
\textbf{MLP-64-deep} architecture was the best performing model, we conducted additional experiments with the AdamW optimizer on this model to explore potential further improvements. All training runs incorporated an early stopping mechanism with a patience of 9 epochs to terminate training when no improvement in validation loss was observed.

\section{Results}
Our results find that our baselines linear regression and logistic regression both did pretty well with accuracies of 0.738 and 0.74, with linear regression having a much lower mean squared error of 0.185 compared to 0.260. 

\subsection{Feature Selection Justification}
\begin{figure}
    \centering
    \includegraphics[width=\linewidth]{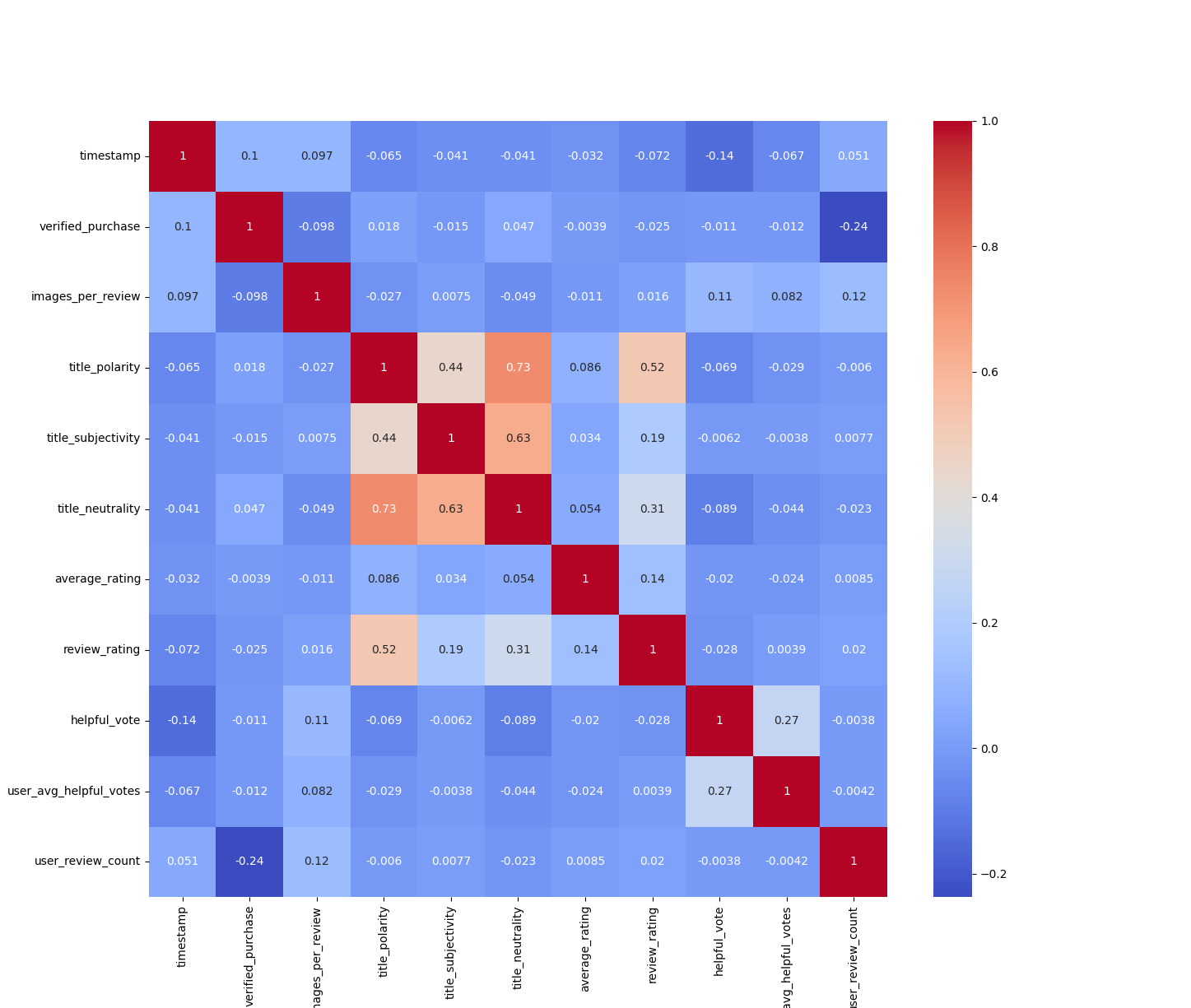}
    \caption{Correlation map based on feature selection}
    \label{fig:correlationmap}
\end{figure}

Initially, we received a result of around 0.73 utilizing our neural network with features such as sentiment taken in. However, after looking at which features worked best, we were able to improve the accuracy to over 90\%. Based on our correlation analysis, we retained several features despite their moderate correlations on the correlation map which can be seen in Figure \ref{fig:correlationmap}. The justification for each feature is as follows:
\subsubsection{User Average Helpful Votes}
The user avg\_helpful\_votes feature was retained because:
\begin{equation}
r_{helpful,avg\_helpful} = 0.27
\end{equation}
This moderate positive correlation ($r = 0.27$) suggests that:
\begin{itemize}
\item A user's historical performance in generating helpful reviews may predict future review helpfulness
\item This feature captures user-specific patterns in review quality
\item It provides a form of user credibility scoring
\item This feature is analogous to our use of similarity metrics earlier in the class
\end{itemize}
\subsubsection{Images per Review}
The images\_per\_review feature was maintained despite its weak correlation:
\begin{equation}
r_{helpful,images} = 0.11
\end{equation}
The justification includes:
\begin{itemize}
\item Visual content often enhances review informativeness
\item The weak positive correlation still indicates some predictive value
\item Based on personal experience, reviews with images are more helpful
\end{itemize}
\subsubsection{Timestamp}
The timestamp feature was preserved for showing negative correlation:
\begin{equation}
r_{helpful,timestamp} = -0.14
\end{equation}
This decision was based on:
\begin{itemize}
\item Temporal aspects may capture review freshness; freshness could impact review order on the page
\item Potential for identifying temporal patterns in review helpfulness
\item Possibility of interaction effects with other features
\end{itemize}
\subsection{Feature Retention Criteria}
Our decision to retain these features was guided by the principle that:
\begin{equation}
|r_{feature,target}| > 0.1 \lor \text{feature provides unique information}
\end{equation}

\begin{table}[htbp]
\caption{Prediction Accuracies and MSE}
\begin{center}
\begin{tabular}{|c|c|c|}
\hline
 & \textbf{Accuracy} & \textbf{MSE} \\
\hline
\textbf{Linear} & 0.7380 & 0.185 \\
\hline
\textbf{Logistic} & 0.7400 & 0.260 \\
\hline
\textbf{MLP-64} & 0.9689 & - \\
\hline
\textbf{MLP-64 Deep} & 0.9690 & - \\
\hline
\textbf{MLP-64 Deep + AdamW} & 0.9691 & - \\
\hline
\textbf{MLP-128} & 0.9682 & - \\
\hline
\textbf{RNN} & 0.7319 & - \\
\hline
\textbf{Transformer} & 0.7319 & - \\
\hline
\end{tabular}
\label{tab:prediction-accuracies}
\end{center}
\end{table}

\subsection{Model Performance}

\subsubsection{Baseline Performance}
Our baselines of linear and logistic regression provided decent predictive accuracy and decently low MSE as seen in Table \ref{tab:prediction-accuracies}. These numbers do beat the RNN and transformer model, which was unexpected. Our MLP models greatly outperformed the sequential models and our baselines, with accuracy nearing 97\% on our test portion of the dataset. It appears that the deeper the model is, the greater the accuracy can be. We don't believe that we ran enough training iterations to over-fit to the dataset, as the dataset is sufficiently large, assuming that there aren't numerous duplicate datapoints.

\subsubsection{Non-Sequential Models Performance}
Each architecture was evaluated using binary classification accuracy metrics with early stopping. The non-sequential models demonstrated superior performance compared to their sequential counterparts.
\begin{itemize}
    \item \textbf{MLP-64}: Achieved 96.89\% accuracy with early stopping at epoch 24, providing an efficient balance between model complexity and performance. The architecture demonstrated optimal convergence characteristics with the standard Adam optimizer.
    
    \item \textbf{MLP-128}: Reached 96.82\% accuracy after 67 epochs, showing comparable performance to MLP-64 but requiring significantly more training iterations.
    
    \item \textbf{MLP-64-deep}: Reached 96.90\%, but obtained the highest accuracy at 96.93\% after 35 epochs, suggesting that the additional depth provided marginal improvements in model performance.

    \item \textbf{MLP-64-deep-adamw}: Reached 96.91\%, achieveing highest accurancy with our early stopping mechanism
\end{itemize}

\subsubsection{Sequential Models Performance}
All sequential models demonstrated consistent performance with binary classification accuracy metrics with early stopping:
\begin{itemize}
    \item \textbf{RNN}: Achieved 73.18\% accuracy with early stopping at epoch 10
    \item \textbf{Transformer}: Attained 73.18\% accuracy with early stopping at epoch 10
\end{itemize}

\subsection{Performance Analysis}
Results reveal a complex relationship between temporal features and model architectures in this classification task. While traditional sequential models (RNNs and Transformers) achieved modest performance (73.18\%), the MLP architecture incorporating temporal features as static inputs significantly outperformed them, reaching 96\% accuracy. This disparity suggests that:

\begin{enumerate}
    \item The nature of temporal relationships in our data may be better suited to static representation rather than sequential processing
    \item The superior performance of the MLP indicates that the relevant temporal patterns might be more effectively captured through direct feature representation rather than through recurrent or attention-based mechanisms
    \item The traditional assumption that sequential models are optimal for temporal data does not hold for this specific classification task
\end{enumerate}

\subsection{Optimization Details}
These results indicate that simpler feed-forward architectures are more suitable for this particular classification task, with the MLP-64-deep model providing the best balance between model complexity and performance.

\section{Conclusion}

Our experimental results revealed significant performance disparities between sequential and non-sequential architectures. The sequential models (RNN and Transformer) failed to surpass even our baseline linear and logistic regression models, which prompts several key observations:

\begin{enumerate}
    \item The temporal features in our dataset demonstrated limited predictive power, ranking only third in our correlation analysis
    \item The complexity of these sequential architectures were too much for our computational and time constraints
\end{enumerate}

Our non-sequential models performed incredibly well, and delivered near-perfect accuracy. From our three models we tested, the MLP-64 Deep performed the best, very slightly edging out the MLP-64 model. It appears that the increase in dimensions internally does not enhance accuracy, and instead decreases it slightly. We believe that these models succeeded due to the simpler nature allowing us to spend more iterations on training, our dataset not having any strong temporal factors, and we chose only relevant features to use in our predictive task.

\nocite{mcauley2022}
\bibliography{bibliography.bib}
\bibliographystyle{plain}
\end{document}